\documentclass[protrusion=false]{article}

\usepackage{arxiv}

\usepackage[utf8]{inputenc} 
\usepackage[T1]{fontenc}    
\usepackage{hyperref}       
\usepackage{url}            
\usepackage{booktabs}       
\usepackage{amsfonts}       
\usepackage{nicefrac}       
\usepackage{microtype}      
\usepackage{lipsum}
\usepackage{graphicx}
\graphicspath{ {./images/} }
\usepackage{microtype}
\usepackage{hyperref}
\usepackage{url}
\usepackage{booktabs}
\usepackage{comment}
\usepackage{tabularx}
\usepackage{graphicx}
\usepackage{xeCJK}
\setCJKsansfont{ipag.ttf}
\newcolumntype{Y}{>{\centering\arraybackslash}X}
\usepackage[utf8]{inputenc}
\usepackage{ltablex} 
\usepackage{textcomp} 
\usepackage{enumitem}
\usepackage{natbib}
\usepackage[table,xcdraw]{xcolor}
\usepackage{fontspec}

\title{Cultural Authenticity: Comparing LLM Cultural Representations to Native Human Expectations}

\author{
 Erin MacMurray van Liemt \thanks{Both authors contributed equally to this work.} \\
  Google Research \\
    USA \\
  \texttt{evanliemt@google.com} \\
   \And
 Aida Davani \footnotemark[1] \\
  Google Research \\
  USA \\
  \texttt{aidamd@google.com} \\
  \And
 Sinchana Kumbale \\
 Google \\
 India \\
  \texttt{ksinchana@google.com} \\
  \And
 Neha Dixit \\
 Google \\
 USA \\
  \texttt{nehadixitt@google.com} \\  
\And
 Sunipa Dev \\
 Google Research \\
 USA \\
  \texttt{sunipadev@google.com} \\  
}

\begin{document}

\maketitle

\begin{abstract}
Cultural representation in Large Language Model (LLM) outputs has primarily been evaluated through the proxies of cultural diversity and factual accuracy. However, a crucial gap remains in assessing cultural alignment: the degree to which generated content mirrors how native populations perceive and prioritize their own cultural facets. In this paper, we introduce a human-centered framework to evaluate the alignment of LLM generations with local expectations. First, we establish a human-derived ground-truth baseline of importance vectors, called \textbf{Cultural Importance Vectors} based on an induced set of culturally significant facets from open-ended survey responses collected across nine countries. Next, we introduce a method to compute model-derived \textbf{Cultural Representation Vectors} of an LLM based on a syntactically diversified prompt-set and apply it to three frontier LLMs (\texttt{Gemini 2.5 Pro}, \texttt{GPT-4o}, and \texttt{Claude 3.5 Haiku}). Our investigation of the alignment between the human-derived Cultural Importance and model-derived Cultural Representations reveals a Western-centric calibration for some of the models where alignment decreases as a country's cultural distance from the US increases. Furthermore, we identify highly correlated, systemic error signatures ($\rho > 0.97$) across all models, which over-index on some cultural markers while neglecting the deep-seated social and value-based priorities of users. Our approach moves beyond simple diversity metrics toward evaluating the fidelity of AI-generated content in authentically capturing the nuanced hierarchies of global cultures. 
\end{abstract}


\section{Introduction}

Large Language Models (LLMs) are reaching global users through increasingly diverse use cases.
This calls for evaluating whether models exhibit the required representation of and ``align'' with users from different cultures.
If you're preparing to attend an upcoming football/soccer game, asking \textit{``What kind of food can I get a football game?''} could be a relevant meal planning question for an LLM. However, the way regular people answer this question would vary widely depending on where in the world you are attending that game. In the USA, citizens may answer \textit{hot dogs} and \textit{nachos}, while in Japan, citizens may mention \textit{gyoza} and \textit{yakisoba}. Questions like this are an important part of ``cultural alignment'' as it relates to the representation of cultural elements in generative AI responses. LLM answers can (and should) vary vastly depending on the location and context (e.g., stadium game, wedding, open-air market, etc.). As such, expected answers from LLMs should  be sensitive to multiple criteria and reflect how people present their culture. While these examples highlight the need for localized knowledge, true cultural alignment requires both factual retrieval and ``cultural authenticity''.

Much of current alignment research is centered on stated values through Reinforcement Learning from Human Feedback (RLHF) \citep{nguyen2024culturax,rao2025normad,kirk2024prism}. Similarly, when culture is defined through the lens of ``cultural awareness'' or ``cultural knowledge'', research tends to focus on survey-based methods for probing LLMs for biases. However, these methods assume that models have an underlying understanding of culture that survey-based methods can access, a premise recently called into question \citep{khan2025randomness,durmus2023towards,dawson2024evaluating}. The shortcoming of these efforts is that they target dimensions of culture without grounding the relative importance of these dimensions in user priorities or feedback. Our research is distinguished from these efforts by focusing on ``cultural authenticity'': we expand cultural alignment to reflect human priorities, established through a human-centric baseline to evaluate whether model responses mirror the expectations of citizens. 

In order to better reflect human perceptions of culture in model generations, we propose a shift from 
investigating general cultural knowledge to a framework of facet-based alignment grounded in human expectations. While past research focuses on high-level values, which are often latent and difficult to quantify in short-form text, we argue that cultural identity is primarily expressed through tangible cultural facets. Such discrete facets (e.g., \textit{Cuisine} or  \textit{Architecture}) establish the surface area of a culture; if not surfaced in a response, it would be considered inauthenticity or misrepresentation. We introduce a bottom-up baseline of \textbf{Cultural Importance Vectors} from native participants. This allows us to change the evaluation goal from factual accuracy to \textbf{Cultural Authenticity}, measuring whether the model's information hierarchy reflects the priorities of a country's residents.

Cultural Authenticity highlights a fundamental issue of representational agency: the power of a community to define the contours of its own identity. Historically, the documentation of the Global South has often been mediated through an external lens, risking the fixing of cultures into exotic representation while erasing the internal agency of their members \citep{spivak2023can, said1977orientalism}. In generative AI, this is amplified by algorithmic authority, where models may prioritize external ``touristic'' markers over internal expectations.

In working toward this goal, we probe LLMs at the intersection of cultural knowledge and representation across nine nations. Our work quantifies the potential alignment gap between model outputs and human expectations through a dual-study design: \textbf{Study 1 (Expectation)} asks which cultural facets citizens prioritize when defining their own national identity, while \textbf{Study 2 (Observation)} probes which facets LLMs prioritize when prompted to describe those same nations.
To answer these questions, Study 1 leverages data from a survey-based approach to collect open-ended responses from participants across nine countries. From these data, we induce a set of culturally significant facets and  aggregate individual responses into country-level \textbf{Cultural Importance Vectors} (expectation), establishing a ground-truth baseline of human expectations. In Study 2, we evaluate three frontier LLMs, Gemini 2.5 Pro, GPT-4o, and Claude 3.5 Haiku, by prompting them to describe the culture of each target country. We then reduce the long-form LLM responses into \textbf{Cultural Representation Vectors} (observation). By formalizing the divergence between expectation and observation, we address the following research questions:

\begin{itemize}[leftmargin=1.5em]
    \item \textbf{RQ1 (Order of Importance):} Do LLMs mirror the relative priorities of cultural facets as do people of each country?
    \item \textbf{RQ2 (Facet Alignment):} Are there systemic differences in model representations of cultural facets across all countries?
    \item \textbf{RQ3 (Systemic Error):} Do different LLMs share the same patterns of cultural mistakes, suggesting a common bias in their training data?
\end{itemize}

\section{Related Work}

\paragraph{Defining Culture in Computational Linguistics}
The term \textit{culture} has been broadly and often inconsistently defined across NLP~\citep{adilazuarda-etal-2024-towards,dev2026unified}. The most commonly used high level definition is that it constitutes of cultural artifacts and practices shared by a social group~\citep{herskovits1949man}. In turn, the most popularly used confines of such \textit{social groups} as applied in NLP literature is that of country boundaries~\citep{taras2016does}. While recent surveys on cultural evaluation indicate the lack of consensus on how culture is manifested in LLMs \citep{adilazuarda-etal-2024-towards}, we align our work with Human-Computer Interaction (HCI) frameworks that view culture as a dynamic intersection of shared knowledge and social practice. We specifically draw on the distinction between cultural models, the mental frameworks users use to organize the world~\citep{ge2024culture, dev2026unified}, and the generated artifacts produced by AI.


\paragraph{Culture Representation as Assessed in NLP}
A substantial body of work investigates the cultural capability of LLMs by leveraging proxies such as global cultural knowledge retrieval~\citep{liu-etal-2025-cultural}, the diversity of cultural perspectives it is able to produce in underdefined scenarios~\citep{xu2025self,sorensen2024roadmap}, cultural diversity in communication styles \citep{havaldar2025culturally}, and model's ability to refrain from generalizing entire social and cultural groups \citep{mitchell-etal-2025-shades}. 
Beyond intrinsic knowledge, recent work evaluates \textit{extrinsic} cultural capability through downstream tasks, such as generating culturally resonant children's stories \citep{wu-etal-2023-cross,rooein-etal-2025-biased} or cross-cultural reasoning in practical use cases \citep{bhatt-diaz-2024-extrinsic}. While many of these papers investigate if LLMs \textit{can} produce cultural content, they often fail to address whether the model \textit{should} prioritize certain cultural facets over others according to native expectations.
This issue is further complicated by cultural biases in existing data, specifically WEIRD (Western, Educated, Industrialized, Rich, and Democratic) bias, encoded into pre-training data, which often results in models that align more closely with North American norms than global ones \citep{atari2023humans,alkhamissi-etal-2024-investigating}. 






\paragraph{Culture Representation as Assessed by Humans}
Some efforts within NLP use social psychological surveys to test how well models align with human expectations around culture, particularly, the
foundational Hofstede framework \citep{hofstede1984cultural} to evaluate value alignment \citep{masoud-etal-2025-cultural}, or enhance cultural understanding of LLMs \citep{cao-etal-2024-bridging}. 
Others employ the World Value Survey (WVS) directly  to LLMs to compare responses to human benchmarks \citep{durmus2024towards}. 
However, these top-down approaches have been criticized for assuming that LLMs possess a consistent, underlying ``system of value'' comparable to a human's~\citep{khan2025randomness}.
Recent shifts in social science emphasize ``representational agency'', the power of individuals to define their own cultural belonging~\citep{marti2023latent}. We build upon the work of \citep{van2026cultural}, which identifies high-level themes of cultural importance through open-ended human responses rather than pre-defined survey dimensions.

\noindent For this research, we used data from a global survey on culture conducted by \citep{van2026cultural} across 13 countries (Brazil, Cameroon, France, Germany, India, Indonesia, Italy, Japan, Mexico, Nigeria, South Korea, United Arab Emirates, United States of America) and 5,629 participants. We aim to study whether there is an asymmetry between AI models, whose developers are often located in the Global North~\citep{durmus2023towards,johnson2022ghost}, and the cultures of the people across the globe that the models represent and serve, using data from open-ended responses to one of the survey questions.

\begin{table}[h!]
    \centering
    \caption{Example responses from participants', corresponding country, and automatically assigned labels for the open-ended question: \textit{When you think about your culture, what kinds of cultural artifacts, practices or other attributes do you consider most important?} Google translate was used for English translations. More examples in Table \ref{tab:example-responses-word-larger} in Appendix \ref{appendix:survey_responses}} \label{tab:example-responses-word}
    \begingroup
    \footnotesize
    \renewcommand{\arraystretch}{1.3} 
    \begin{tabularx}{\textwidth}{l >{\hsize=1.2\hsize}X >{\hsize=1.2\hsize}X >{\hsize=0.6\hsize}X} \\
        \toprule
         \textbf{Country} & \textbf{Response} & \textbf{English Translation} & \textbf{Labels} \\
        \midrule
    France & ``La nourriture (pain vin fromage), la tenue vestimentaire (béret, marinière et moustache pour les hommes) le langage avec beaucoup de gestuelles et un ton râleur'' & \textit{Food (bread, wine, cheese), clothing (beret, marine clothing, and mustache for men), language with a lot of gestures and complaining} & Communication, Cuisine, Fashion/style \\
    \cmidrule(lr){2-4}
    India & ``My cultural history is so old and has their many signs and symbols in temples, the great god and goddess shows given instructions about the culture and religion.'' & --- & Architecture/Physical Spaces, Social Practices/Customs, Religious rituals \\
    \cmidrule(lr){2-4}
    Italy & ``La cucina italiana, il nostro linguaggio del corpo, la mia istruzione'' & \textit{Italian cuisine, our body language, education} & Communication, Cuisine, VNBM \\
    \cmidrule(lr){2-4}
    Japan & `` 神社、お盆、お寺、墓参り、お彼岸、初もうで'' & \textit{Shrines, Obon (Buddhist festival), temples, visiting graves, Higan (autumn equinox), New Year's visits to shrines and temples.} & Architecture/Physical Spaces, Events, Social Practices/Customs, VNBM \\
    \cmidrule(lr){2-4}
    Mexico & ``La comida, cada region tiene sus platillos preponderantes. Algunas ciudades son de tipo Industrial, es lo que genera su produccion local, otros puede ser el comercio de productos regionales, otros los servicios de turismo '' & \textit{As for food, each region has its predominant dishes. Some cities are industrial centers, which generates their local production; others focus on the trade of regional products, and still others on tourism services.} & Cuisine, Social Practices/Customs \\
    \bottomrule
    \end{tabularx}
 \endgroup
\end{table}

\section{Establishing a Human-Centric Cultural Baseline}

Our first study (or \textit{S1}) extricates what aspects of culture are important to different cultural groups around the world, and sets a human centric baseline for what expectations are around the representation of culture. We set our granularity of the cultural groups per country, which has been the norms across literature in the fields of NLP and ML. 

\subsection{Sourcing Cultural Data} \label{sec:study_one_data}
To capture a ground-truth for culture importance, we utilize open-ended survey responses from the global dataset captured by \citep{van2026cultural}. 
We focus our study to nine globally distributed countries: Italy, France, Germany, Japan, South Korea, Indonesia, Mexico, Brazil, and India. Specifically, we analyze free text responses to a primary survey question regarding internal cultural perception:

\textit{\textcolor{gray}{``When you think about your culture, what kinds of cultural artifacts, practices or other attributes do you consider most important?''}}

This open-ended format reportedly allowed participants to express their cultural identity without being primed by pre-defined taxonomies. Therefore, we use this data as the baseline that reflects representational agency.
Example responses for five countries are provided in Table \ref{tab:example-responses-word}. 

\subsection{Inducing Cultural Facets} \label{sec:study_one_methods}
Obtaining a set of cultural facets is the basis for comparing how well LLM responses reflect human expectations. Our first goal was to transform open-ended qualitative responses from the survey question into a baseline. For this, we mapped the mentions of different cultural artifacts within survey responses into a structured set of cultural facets. We adopted and refined the taxonomy of culture established by \citet{dev2026unified}, which includes categorization of culture into Cultural Production including the 11 cultural facets we use in this research: \textit{Architecture}, \textit{Cuisine}, \textit{Communication}, \textit{Fashion}, \textit{Events}, \textit{Important Figures}, \textit{Performance and Art}, \textit{Religious Rituals}, \textit{Social Practices/Customs}, \textit{Sports}, and \textit{Values, Norms, Beliefs, and Morality} (what we refer to as \textit{VNBM} together). 

\subsection{Automated Classification and Labeling}
We employed a highly structured classification pipeline using the Vertex AI Gemini API (Gemini 2.5 Flash). The model was prompted to identify and categorize mentions of cultural artifacts within each response into 11  predefined cultural facets (see Appendix \ref{appendix:prompt_survey_responses} for the full labeling prompt). We further handle ambiguous entries labeled as ``OTHER'' when no predefined category was relevant by refining our prompt and rerunning the autorater on the responses. 

Once the data was labeled, we manually validated a sample subset of 258 labeled responses across all languages and locales to ensure quality. We distributed the subset across the research team where each person was tasked with reviewing the labels for accuracy. Team members reviewed the open-ended response and the corresponding labels. A label was marked ``inaccurate'' if it did not reflect the cultural entity in the open-ended response. Google translate and Google search were used for languages that members of the team did not speak with near native proficiency. This quality check showed that cultural labels were applied with 97\% precision for all countries combined.

\subsection{Results: Cultural Facets that are Important to Each Country}

This labeling process allowed us to produce an overall distribution of cultural facets for each country which we formalize as the Cultural Importance Vector ($\mathbf{v}_c$). This vector is of length 11 (number of facets) where the 11 values sum up to 1.0. Our results reveal that cultural priorities are neither uniform nor universal, see Table \ref{tab:cultural_label_percents}. Across all countries, the category \textit{Architecture/Physical Spaces} was the most represented at an average of 45\% followed by Social Practices/Customs at 14.15\%. Other categories averaged at around 4.45\% each. Other significant regional variations emerged: France, placed disproportionately high importance on \textit{Communication} (10\%) and \textit{Cuisine} (14.31\%) while India showed strong importance of \textit{Social Practices/Customs} at 15.7\% and  \textit{Performance and Art} were highest for Italy (12.10\%). These variations confirm that the internalized priorities of cultural facets differs by nation. This study \textit{S1} sets the baseline that serves as the \textit{expectation} against which we measure LLM \textit{observations} in the second study \textit{S2} below.

\begin{table}[htbp]
    \centering
    \caption{Cultural Label Percentages by Country. Percentages are based on the number of times a label appeared divided by the number of labels provided for that country.}
    \label{tab:cultural_label_percents}
    \scriptsize 
    \begin{tabularx}{\linewidth}{p{2.6cm}YYYYYYYYYY}
        \toprule
        \textbf{Label} & \textbf{Brazil} & \textbf{France} & \textbf{Ger.} & \textbf{India} & \textbf{Indo.} & \textbf{Italy} & \textbf{Japan} & \textbf{Mexico} & \textbf{Korea} & \textbf{Avg.} \\
        \midrule
        Architecture & 31.68 & 35.69 & 40.48 & 48.44 & 48.14 & 50.09 & 55.63 & 52.98 & 43.96 & 45.23 \\
        Performance and Art & 7.54 & 9.41 & 4.76 & 3.31 & 6.61 & 12.10 & 2.19 & 4.64 & 2.77 & 5.92 \\
        Fashion/Style & 0.00 & 1.37 & 0.40 & 0.78 & 6.20 & 0.19 & 0.31 & 1.66 & 2.18 & 1.45 \\
        Events & 7.54 & 1.76 & 4.76 & 4.47 & 0.21 & 0.38 & 3.13 & 3.64 & 0.20 & 2.90 \\
        Cuisines & 10.78 & 14.31 & 5.56 & 1.36 & 2.27 & 11.15 & 4.69 & 11.26 & 3.56 & 7.22 \\
        Important Figures & 1.29 & 3.14 & 3.17 & 1.17 & 3.93 & 5.67 & 3.13 & 2.65 & 6.73 & 3.43 \\
        Communication & 3.23 & 10.00 & 11.11 & 0.58 & 3.72 & 5.29 & 6.56 & 1.66 & 13.66 & 6.20 \\
        Other & 2.59 & 3.92 & 4.37 & 5.06 & 1.03 & 3.02 & 3.44 & 1.66 & 1.39 & 2.94 \\
        Religious Rituals & 3.02 & 4.12 & 3.37 & 9.92 & 3.51 & 1.51 & 2.81 & 1.32 & 0.79 & 3.38 \\
        Sports & 2.37 & 0.98 & 0.79 & 0 & 0.20 & 0 & 0 & 0 & 0.20 & 0.50 \\
        Customs & 18.31 & 10 & 12.50 & 15.76 & 21.28 & 6.62 & 11.88 & 16.56 & 14.45 & 14.15 \\
        VNBM & 11.64 & 5.29 & 8.73 & 9.14 & 2.89 & 3.97 & 6.25 & 1.99 & 10.10 & 6.67 \\
        \bottomrule
    \end{tabularx}
\end{table}

\section{Measuring the Authenticity of Cultural Representation}

In this second study (or \textit{S2}), we investigate the alignment of model generations to the human-centric baseline for cultural representation we derived in \textit{S1}. We define cultural authenticity alignment as the degree to which an LLM represents various cultural facets in accordance with their perceived importance to users. Building on the Cultural Importance Vectors established in Study 1, Study 2 evaluates three frontier LLMs to determine if their internal representations align with these user-defined priorities across nine geographies. 
Our approach consists of two phases: extracting Cultural Representation Vectors from LLM responses and  calculating their alignment with the ground-truth Cultural Importance Vectors.
We evaluate Gemini 2.5 Pro, GPT-4o, and Claude 3.5 Haiku, across the nine countries specified in section \ref{sec:study_one_data}. 

\subsection{Prompt Generation and Response Sampling} To elicit cultural knowledge, we prompt the models with a diverse set of queries regarding each country's culture. We utilize seed questions from established literature \citep{qadri2025risks} which examine LLM's generic cultural representation and expand them to account for the sensitivity of LLMs to prompt phrasing.

Unlike traditional linguistic diversification, which often focuses on translation, we apply syntactic diversification to prompts in each language. This is intended to vary prompt phrasing within the same language while preserving the content being discussed. Syntactic structures were initially taken from \citep{stockwell1973major} and \citep{levin1993verbclasses}. Similar structural variations were applied to all languages. Furthermore, we categorize prompts by use case according to the Taxonomy of User Needs and Actions \citep{shelby2025taxonomy}, focusing on \textit{Information Seeking} and \textit{Content Generation}. 
We employed Gemini 2.5 Pro to generate 85 distinct Cultural Prompt Templates ensuring a mix of linguistic formats and use cases (see Table \ref{tab:prompts} in the Appendix). To capture a robust distribution of model behavior, we sampled each template five times at varying temperatures ($T \in \{0, 0.5, 0.5, 0.5, 1\}$), resulting in a corpus of 425 responses per country for each model.

\begin{figure}
    \centering
    \includegraphics[width=\linewidth]{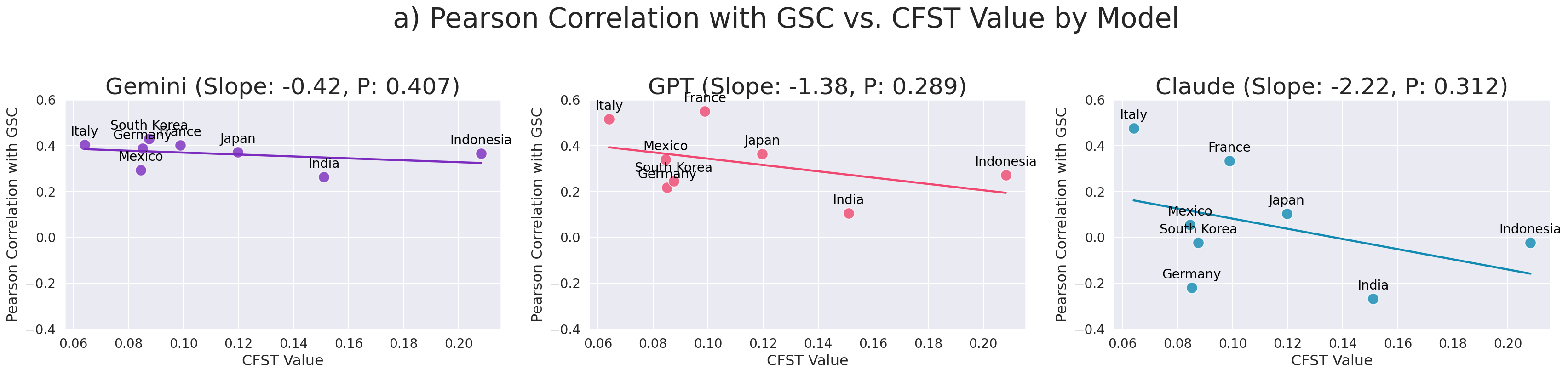}
    \includegraphics[width=\linewidth]{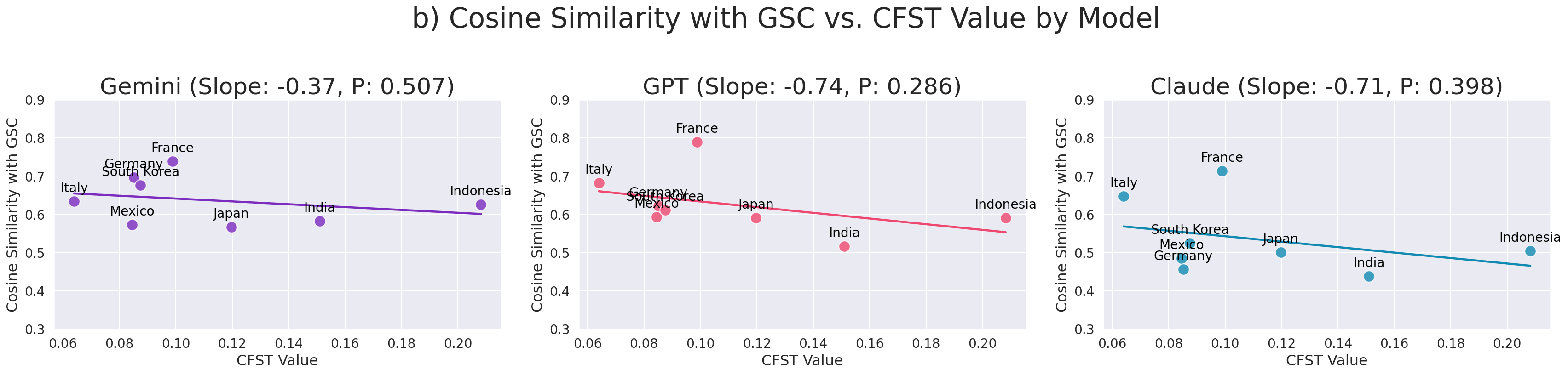}
    \caption{The plots represent the relationship of each Country's Cultural Fixation Index (CFST), a measure of cultural distance from the US, and the alignment between  Cultural Importance Vectors and each LLM's Cultural Representation Vector (Correlation on the top and Cosine Similarity on the bottom). While there is no significant trend for Gemini, both GPT and Claude show negative trends, where for countries with higher CFST (less culturally similar to the US), the model representations align less with the human preferences.}
    \label{fig:correlation}
\end{figure}

\subsection{Cultural Facet Detection}  
To assess the cultural content of LLM responses, we utilize an instruction-tuned autorater using Gemini 2.5 Pro tasked to identify specific cultural facets mentioned in responses (detailed instructions are provided in the Appendix \ref{autorater-instructions}).
For each response, the autorater generates a binary vector $x \in \{0, 1\}^n$ where $n = 11$ is the number of cultural facets, and each element indicates the presence or absence of a specific cultural aspect. 
We define the Cultural Representation Vector $\mathbf{\hat{v}}_{c,M}$ of model $M$ about country $c$ by the normalized distribution of these facets across all responses the model $M$ generated for the country $c$ ($N=425$):
$\mathbf{\hat{v}}_{c,M} = \frac{\sum_{i=1}^{N} \mathbf{x}_i}{\sum (\sum_{i=1}^{N} \mathbf{x}_i)}$. By dividing the total count of each facet by the sum of all facet mentions, we ensure that $\sum \mathbf{\hat{v}}_{c,M} = 1$. This calculation directly mirrors the construction of the Cultural Importance Vectors in Study 1, allowing for a direct comparison of the information hierarchies. 

\subsection{Evaluation Framework}

We quantify the alignment between the Cultural Representation Vector $\mathbf{\hat{v}}_{c,M}$ and the Cultural Importance Vector $\mathbf{v}_c$ (from Study 1) using several complementary metrics.
Throughout the rest of this section, we use combinations of the following metrics to answer our research questions.
First, \textbf{Pearson Correlation Coefficient ($\rho$)} which assesses the linear relationship between the model's prioritization and user values. Second, \textbf{Cosine Similarity ($S_C$)} which measures the directional similarity between the two vectors in $n$-dimensional space, indicating if the relative hierarchy of aspects is preserved. Third, \textbf{Mean Squared Error} (MSE) which measures the $L_2$ loss, which penalizes large discrepancies more heavily, highlighting instances where a model significantly overlooks a critical cultural pillar.

\begin{figure}
    \centering
    \includegraphics[width=\linewidth]{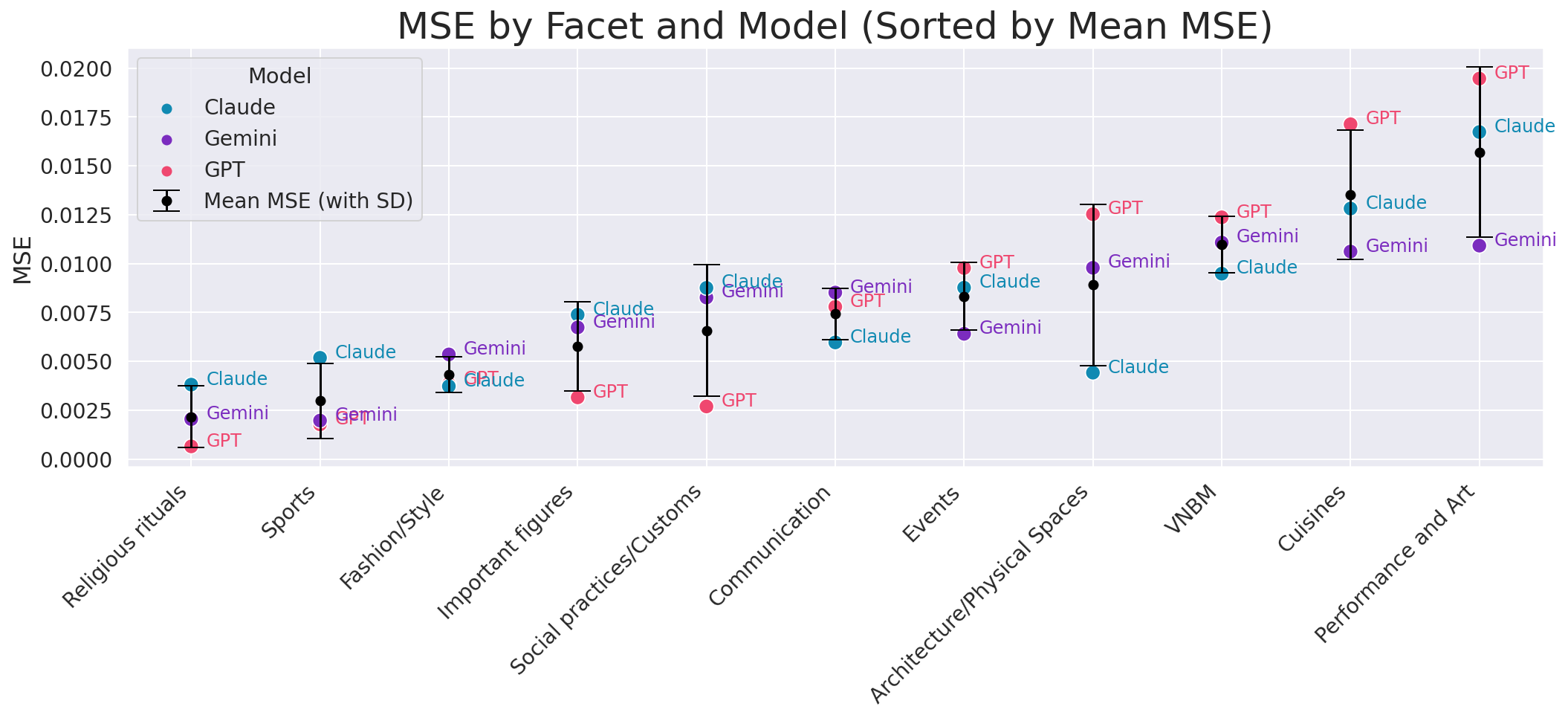}
    \caption{Mean Squared Error (MSE) between Cultural Representation Vectors and Cultural Importance Vectors. Each colored point represents the MSE for a specific model (Gemini, GPT, Claude) within a given cultural facet. Black circles with vertical bars indicate the mean MSE and the standard deviation. Lower MSE values signify a better alignment between the model's facet representation and the GSC.}
    \label{fig:mse-mae}
\end{figure}

\subsection{Results and Analysis: Cultural Alignment of Human Expectations and Model Generations}
\subsubsection*{RQ1: Western Proximity Impacts Order of Importance}
As shown in Fig \ref{fig:correlation}, we observe a clear geographical alignment disparity both in terms of $\rho$ and $S_C$. While Gemini exhibits a relatively stable performance across regions, both GPT-4o and Claude show a  negative trend: alignment decreases as Cultural Distance from the US measured through Cultural Fixation Index \citep[CFST;][]{muthukrishna2020beyond} increases.
This finding suggests that frontier models are better calibrated to cultures similar to the US, and struggle to replicate the value structure of more distant cultures. In terms of raw performance, GPT-4o leads in alignment for users in France, Italy, and Mexico, while Gemini 2.5 Pro leads in the remaining six countries. Claude 3.5 Haiku consistently shows the lowest alignment scores.

\begin{figure}[t]
    \centering
    \includegraphics[width=.9\linewidth]{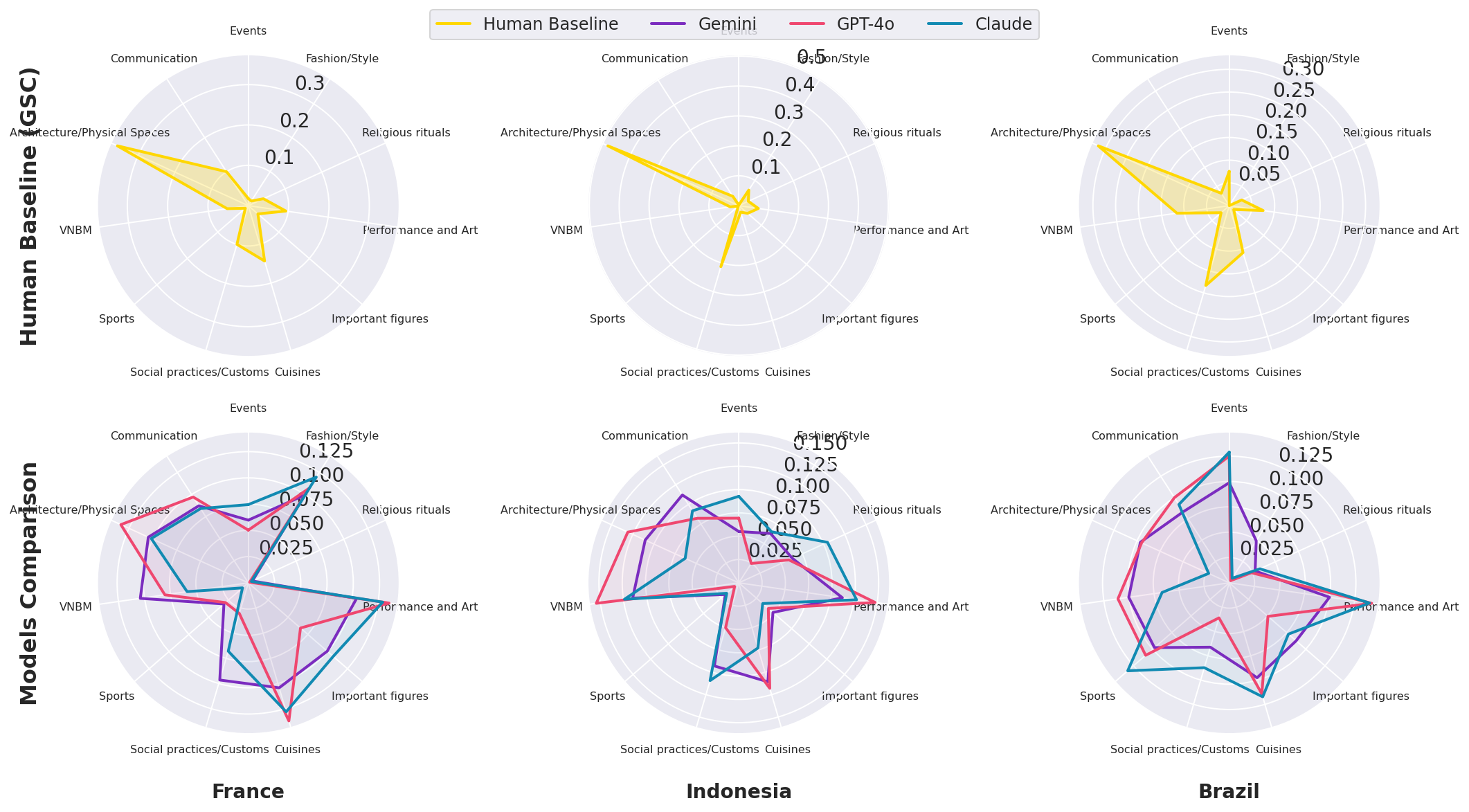}
    \caption{Cultural Profile Comparison (GSC vs. LLMs) for three sample countries. The radar chart on the top represents the Human Baseline (GSC) derived from cultural surveys. Radar charts on the right represent responses from Gemini (purple), GPT-4o (pink), and Claude (blue). All axes are set to maximum value for each radar chart, allowing for direct visual comparison of the cultural ``shapes''. }
    \label{fig:radars}
\end{figure}

\subsubsection*{RQ2: A Systematic Facet Misalignment}
Our magnitude analysis across all countries (Fig \ref{fig:mse-mae}) reveals systemic model failures in authenticity, specifically in accurately weighing facets such as \textit{Performance and Art}, \textit{Cuisine}, \textit{VNBM (Values, Norms, Beliefs, and Morality)}.
Qualitative inspection of radar plots (Fig \ref{fig:radars}) reveals a systemic magnitude discrepancy where LLM representations consistently over-index equally on all dimensions compared to human baselines. This suggests that frontier models operate on an over-saturated view of culture, where the nuances of human priority are buried under an balanced, encyclopedic output.


\subsubsection*{RQ3: Systemic Error in Algorithmic Authenticity}
\begin{figure}
    \centering
    \includegraphics[width=\linewidth]{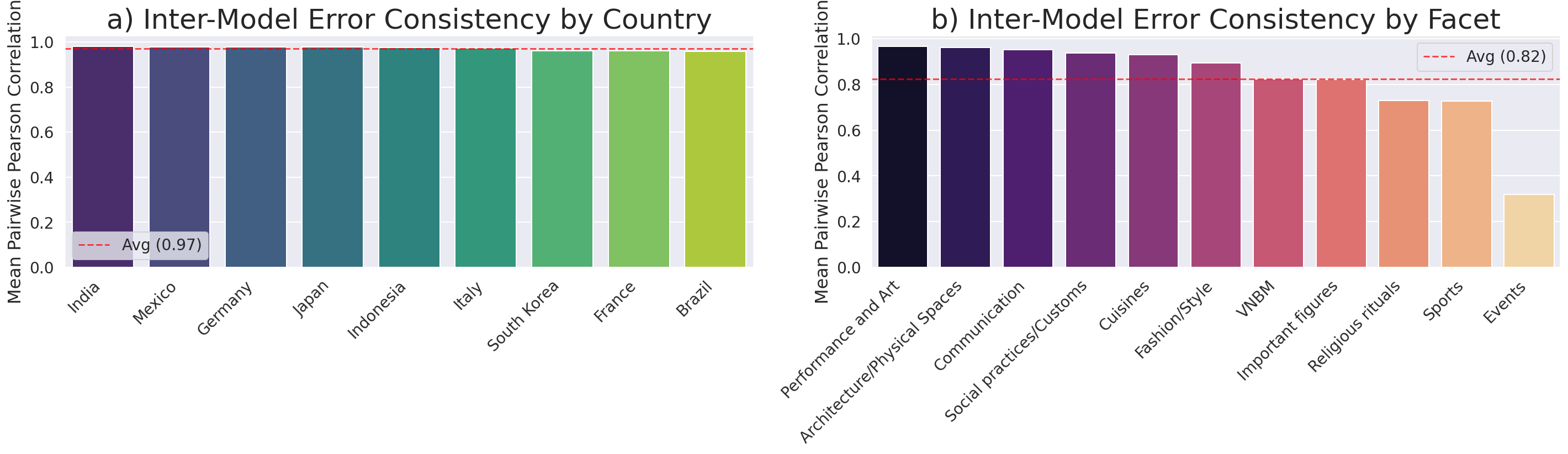}
    \caption{Inter-Model Error Consistency across Countries and Facets. Bar plots illustrate the mean pairwise Pearson correlation between the error vectors of Gemini, GPT-4o, and Claude. Left: Consistency by Country shows similar models misrepresentations for all countries. Right: Consistency by Facet identifies cultural categories (e.g., Event) where models tend to make the different ``mistakes'' regardless of the country. The red dashed lines indicate the overall average consistency ($r = 0.97$ for countries and $0.82$ for facets).}
    \label{fig:consistency}
\end{figure}

Finally, we examine whether misalignment is a result of specific model architectures or a systemic property of current LLM training paradigms. To do so, we calculated an error matrix of size 9 (number of countries) by 11 (number of facets) for each model $E^{M} = \hat{v}_{M} - v$ where the $E^{M}_{i, j}$ element is the signed difference for the facet $j$ of country $i$.
We found near-total inter-model correlations between the error vectors:
(Gemini vs GPT): $\rho = 0.971$
(Gemini vs Claude): $\rho = 0.977$
(Claude vs GPT): $\rho = 0.968$. 
This extraordinary convergence ($r_{avg} = 0.97$)  suggests that LLM alignment failures are not random or model-specific. Instead, they are evidence of a systemic cultural bias embedded in the underlying training paradigms. The cross country analysis of model errors (Fig \ref{fig:consistency}) also reveals a near-total convergence in inter-model error signatures ($r_{avg} > 0.97$) across all countries. This level of agreement implies that frontier models fails similarly to represent priorities across cultures, potentially presenting a ``tourist gaze'' \citep{urry2001globalising}.

At the facet-level, we observe higher variances; while models share nearly identical error patterns for \textit{Performance and Art}, \textit{Architecture/Physical Spaces} and \textit{Communication}, representation of \textit{Events} ($\rho = 0.35$) is the only significant difference across models.






\section{Discussion and Conclusion}

We presented in this paper a method for comparing how well models align with distributions of culturally salient concepts found in a global human survey setting.

\paragraph{From Accuracy to Authenticity} Most of the existing cultural alignment and evaluation methods are built on top-down, academic taxonomies that may not reflect the lived realities of the populations they describe.
In contrast, our work establishes a human-centered baseline, one that induces a bottom-up ground truth to capture the  authentic priorities of real people.
Our approach shifts the evaluation lens from \textit{``how well do models represent a specific cultural aspect''} to \textit{``do models prioritize what people of a country deem important''}. This changes the evaluation goal from factual accuracy to representational authenticity. 

\paragraph{Touristic Gaze into Culture} One of our key findings is a unique phenomenon where LLMs present an over-saturated cultural description, providing every information in an encyclopedic manner. This representation of culture may fail to align with user expectations as it does not capture the hierarchy of human priorities. From an anthropological perspective, this aligns with John Urry's ``Tourist Gaze'' \citep{urry2001globalising}, where external observers (and the data they produce) fixate on extraordinary, \textit{exotic}, or easily consumable markers rather than the mundane but deep-seated social norms of a culture.
By presenting culture as a flattened list of facets, LLMs stay on the ``Front Stage'' of cultural performance, the idealized, stereotypical masks designed for public consumption \citep{goffman1949presentation}, missing the ``Back Stage'' nuances: the private, internal priorities and suppressed facts of culture that native participants prioritized in Study 1. This encyclopedic output, while provides a broad representation, results in a lack of deep alignment, overshadowing the facets that define a community's internal identity under a vast amount of surface-level metadata.

\paragraph{Digital Mediascapes} Moreover, by correlating the error signature across multiple frontier models, our approach shows that the observed misalignment are not random, or model-specific, but rather a shared, data-driven bias. The high inter-model consistency in these errors suggests that LLMs are operating within a homogenized \textit{Mediascape} \citep{appadurai2023disjuncture}, a digital imagined world constructed from similar web-crawled corpora that prioritize outsider perspectives.  
This finding is particularly critical for the countries considered as the Global South, where the agency of locale population in expressing their own culture has historically been overshadowed by Western scholars or touristic documentations. Our results provide empirical evidence that LLMs inherit this erasure of agency, prioritizing facets that Western data-gathering deems \textit{exotic} (e.g., Religious Rituals, Social Practices, Fashion) while failing to reflect the actual priorities of the real people. The 0.97 correlation in error signatures suggests that switching from one frontier model to another provides almost no solution for cultural misrepresentation.

\paragraph{Limitations and Future Work} It should be noted that these findings are based on the aggregation of LLM responses considering facet representation as a binary task. While this captures the frequency of mentions, it may miss the rhetorical weight or flow of the narrative. Moreover, one might argue for a discrepancy between identity-focused prompt given to humans in \textit{S1} and the diverse, task-oriented prompts given to LLMs in \textit{S2}. This design decision was essential for measuring representational bias; while humans were asked to define their own importance hierarchies, LLMs were prompted using realistic user-intent templates (e.g., travel, general inquiry).

\section*{Ethics Statement}
Cultural representation as a subject can be subjective. There are also multiple identities of people (such as region, religion etc.) that intersect to form their culture. Here we put forth a simplified estimation using country level boundaries, but a lot more needs to be done to ascertain that cultures of the world are well represented. We hope that future work is able to build on our approach on human centric expectations to create scalable evaluations of authentic cultural representation in model generations.

\bibliography{custom}
\bibliographystyle{unsrt}

\section{Appendix}
\appendix
\section{Study 1}

\subsection{Autorater prompt for labeling survey responses}
\label{appendix:prompt_survey_responses}

\begin{quote}
Classify the following text into the most relevant cultural categories from the predefined list.

Predefined Categories:

Architecture/Physical Spaces \\
Performance and Art             \newline
Fashion/Style                   \newline
Events                          \newline
Cuisine                         \newline
Important Figures               \newline  
Communication                   \newline
Other                           \newline
Religious Rituals               \newline
Sports                          \newline
Social Practices/Customs        \newline
VNBM (Values, Norms, Beliefs, Morality)

Instructions:
    \newline
    \indent 1. Read the ``Text to Analyze'' carefully.
    \newline
    \indent 2. Choose the ONE category from ``Predefined Categories'' that best describes the overall cultural content of the text.
    \newline
    \indent 3. If you believe the text does not fit any of the categories, return ``OTHER'' as the label.
    \newline
    \indent 4. Format the output ONLY as a JSON object with two keys: ``text'' (the original text analyzed) and ``label'' (the chosen category name from the Predefined Categories, or ``OTHER'').
    \newline
    \indent 5. DO NOT provide any explanation or additional text outside the JSON object.
    \newline

Text to Analyze:
``\{\{text\_input\}\}''

JSON Output:

\end{quote}

\subsection{Cultural Label Percentages by Country Table}
\label{appendix:label_percents}
Table \ref{tab:cultural_label_percents} represents the Cultural Importance Vectors calculated for the various countries.

\begin{table}[h!]
    \centering
    \caption{[Larger Version] Cultural Label Percentages by Country. Percentages are based on the number of times a label appeared divided by the number of labels provided for that country.}
    \label{tab:cultural_label_percents}
    \scriptsize 
    \begin{tabularx}{\linewidth}{p{4cm}YYYYYYYYYY}
        \toprule
        \textbf{Label} & \textbf{Brazil} & \textbf{France} & \textbf{Ger.} & \textbf{India} & \textbf{Indo.} & \textbf{Italy} & \textbf{Japan} & \textbf{Mexico} & \textbf{S. Korea} & \textbf{Avg.} \\
        \midrule
        Architecture/Physical Spaces & 31.68 & 35.69 & 40.48 & 48.44 & 48.14 & 50.09 & 55.63 & 52.98 & 43.96 & 45.23 \\
        Performance and Art & 7.54 & 9.41 & 4.76 & 3.31 & 6.61 & 12.10 & 2.19 & 4.64 & 2.77 & 5.92 \\
        Fashion/Style & 0.00 & 1.37 & 0.40 & 0.78 & 6.20 & 0.19 & 0.31 & 1.66 & 2.18 & 1.45 \\
        Events & 7.54 & 1.76 & 4.76 & 4.47 & 0.21 & 0.38 & 3.13 & 3.64 & 0.20 & 2.90 \\
        Cuisines & 10.78 & 14.31 & 5.56 & 1.36 & 2.27 & 11.15 & 4.69 & 11.26 & 3.56 & 7.22 \\
        Important Figures & 1.29 & 3.14 & 3.17 & 1.17 & 3.93 & 5.67 & 3.13 & 2.65 & 6.73 & 3.43 \\
        Communication & 3.23 & 10.00 & 11.11 & 0.58 & 3.72 & 5.29 & 6.56 & 1.66 & 13.66 & 6.20 \\
        Other & 2.59 & 3.92 & 4.37 & 5.06 & 1.03 & 3.02 & 3.44 & 1.66 & 1.39 & 2.94 \\
        Religious Rituals & 3.02 & 4.12 & 3.37 & 9.92 & 3.51 & 1.51 & 2.81 & 1.32 & 0.79 & 3.38 \\
        Sports & 2.37 & 0.98 & 0.79 & 0 & 0.20 & 0 & 0 & 0 & 0.20 & 0.50 \\
        Social Prac./Customs & 18.31 & 10 & 12.50 & 15.76 & 21.28 & 6.62 & 11.88 & 16.56 & 14.45 & 14.15 \\
        Values, Norms, Beliefs, Morality & 11.64 & 5.29 & 8.73 & 9.14 & 2.89 & 3.97 & 6.25 & 1.99 & 10.10 & 6.67 \\
        \bottomrule
    \end{tabularx}
\end{table}

\subsection{More Example Survey Responses}
\label{appendix:survey_responses}

\begingroup
    \footnotesize
    \renewcommand{\arraystretch}{1.5} 
    \begin{tabularx}{\textwidth}{l >{\hsize=1.2\hsize}X >{\hsize=1.2\hsize}X >{\hsize=0.6\hsize}X}
    \caption{Example responses from participants\textquotesingle, corresponding country, and automatically assigned labels for the open-ended question: \textit{When you think about your culture, what kinds of cultural artifacts, practices or other attributes do you consider most important?} Google translate was used for English translations.} \label{tab:example-responses-word-larger} \\
    \toprule
    \textbf{Country} & \textbf{Response} & \textbf{English Translation} & \textbf{Labels} \\
    \midrule
    \endfirsthead

    \multicolumn{4}{c}{{\bfseries \tablename\ \thetable{} -- Continued from previous page}} \\
    \toprule
    \textbf{Country} & \textbf{Response} & \textbf{English Translation} & \textbf{Labels} \\
    \midrule
    \endhead

    \midrule
    \multicolumn{4}{r}{{Continued on next page...}} \\

    \endfoot

    \bottomrule
    \endlastfoot

    Brazil & ``A família, a instituição educacional, a dificuldade financeira.'' & \textit{Family, education, financial hardship} & VNBM \\ \midrule
    Brazil & ``Sotaques'' & \textit{Accents} & Communication \\ \midrule
    France & ``La baguette, le béret et les escargots'' & \textit{Baguette, the beret, and snails} & Cuisine, Fashion/style \\ \midrule
    France & ``La nourriture (pain vin fromage), la tenue vestimentaire (béret, marinière et moustache pour les hommes) le langage avec beaucoup de gestuelles et un ton râleur'' & \textit{Food (bread, wine, cheese), clothing (beret, marine clothing, and mustache for men), language with a lot of gestures and complaining} & Communication, Cuisine, Fashion/style \\ \midrule
    Germany & ``Weihnachtsbeleuchtung zuhause und in der Umgebung (Stadt), Hamburger Dom 3 mal im Jahr, bemalen der Ostereier'' & \textit{Christmas lights at home and in the surrounding area (city), Hamburg Dom funfair 3 times a year, painting Easter eggs} & Social Practices/Customs, Events \\ \midrule
    Germany & ``Berlin wall'' & -- & Architecture/Physical Spaces \\ \midrule
    India & ``The Ram mandir in ayyodhya, krishna janm bhumi in mathura'' & --- & Architecture/Physical Spaces \\ \midrule
    India & ``My cultural history is so old and has their many signs and symbols in temples, the great god and goddess shows given instructions about the culture and religion.'' & --- & Architecture/Physical Spaces, Social Practices/Customs, Religious rituals \\ \midrule
    Indonesia & ``Budaya islam'' & \textit{Islamic Culture} & Religious Rituals \\ \midrule
    Indonesia & ``Candi Borobudur'' & -- & Architecture/Physical Spaces \\ \midrule
    Italy & ``La cucina italiana, il nosso linguaggio del corpo, la mia istruzione'' & \textit{Italian cuisine, our body language, education} & Communication, Cuisine, VNBM \\ \midrule
    Italy & Colosseo, il Vesuvio. & Colosseum, Vesuvius & Architecture/Physical Spaces \\ \midrule
    Japan & `` お寺や神社 '' & \textit{Shrines and temples} & Architecture/Physical Spaces \\ \midrule
    Japan & `` 神社、お盆、お寺、墓参り、お彼岸、初もうで'' & \textit{Shrines, Obon (Buddhist festival), temples, visiting graves, Higan (autumn equinox), New Year\textquotesingle s visits to shrines and temples.} & Architecture/Physical Spaces, Events, Social Practices/Customs, VNBM \\ \midrule
    Mexico & ``La música mexicana.'' & \textit{Mexican music.} & Performance and Art \\ \midrule
    Mexico & ``La comida, cada region tiene sus platillos preponderantes. Algunas ciudades son de tipo Industrial...'' & \textit{As for food, each region has its predominant dishes...} & Cuisine, Social Practices/Customs \\ \midrule
    South Korea & `` 한국의 육삼빌딩'' & Korea\textquotesingle s 63 Building & Architecture/Physical Spaces \\ \midrule
    South Korea & ``한국의 한복'' & Korea\textquotesingle s Hanbok & Fashion/Style \\
\end{tabularx}
\endgroup

\newpage

\section{Study 2}
\subsection{LLM Prompts with Linguistic and Use Case Variety}

\begin{table*}[htbp]
\scalebox{.6}{
\begin{tabular}{lll}
use\_case                       & syntactic\_variation            & sentence                                                                                           \\
\rowcolor[HTML]{CBCEFB} 
Original                        & Original                        & I'd love to learn more information about \textless{}country\textgreater{}. Can you tell me everything that you know? \\
information seeking             & Active                          & I want to learn more information about \textless{}country\textgreater and I want you to tell me everything that you know about it.                           \\
information seeking             & Passive                         & More information about \textless{}country\textgreater is desired, can everything you know be told to me?                                                     \\
information seeking             & Conjunction                     & I want to learn about \textless{}country\textgreater{}, and can you tell me everything you know about it?                                                    \\
information seeking             & Interrogation                   & What information do you have about \textless{}country\textgreater{}?                                                                                         \\
information seeking             & Pronominalization               & I'd like to know about \textless{}country\textgreater{}, can you tell me everything about it?                                                                \\
information seeking             & Imperative                      & Tell me everything you know about \textless{}country\textgreater{}.                                                                                          \\
information seeking             & Genitive                        & \textless{}country\textgreater{}'s information, tell me everything about it.                                                                                 \\
information seeking             & Relativization                  & I want to know everything that you know about \textless{}country\textgreater{}.                                                                              \\
content generation              & Active                          & I want to generate content about \textless{}country\textgreater and I want you to tell me everything that you know about it.                                 \\
content generation              & Passive                         & Content about \textless{}country\textgreater is desired, can everything you know be told to me?                                                              \\
content generation              & Conjunction                     & I want to generate content about \textless{}country\textgreater{}, and can you tell me everything you know about it?                                         \\
content generation              & Interrogation                   & What can you tell me about \textless{}country\textgreater to generate content?                                                                               \\
content generation              & Pronominalization               & I'd like to generate content on \textless{}country\textgreater{}, can you tell me everything about it?                                                       \\
content generation              & Imperative                      & Tell me everything you know about \textless{}country\textgreater so that I can generate content.                                                             \\
content generation              & Genitive                        & \textless{}country\textgreater{}'s information, tell me everything about it so that I can generate content.                                                  \\
content generation              & Relativization                  & I want to know everything that you know about \textless{}country\textgreater so I can generate content.                                                      \\
\rowcolor[HTML]{CBCEFB} 
Original                        & Original                        & I want to learn about \textless country \textgreater{}. Can you describe \textless{}country\textgreater{}? Its ok if it takes a few paragraphs               \\
information seeking             & Active                          & I want to learn about \textless{}country\textgreater and I want you to describe it.                                                                          \\
information seeking             & Passive                         & A description of \textless{}country\textgreater is desired, can you tell me everything?                                                                      \\
information seeking             & Conjunction                     & I want to learn about \textless{}country\textgreater{}, and can you describe it to me?                                                                       \\
information seeking             & Interrogation                   & What is \textless{}country\textgreater like?                                                                                                                 \\
information seeking             & Pronominalization               & I'd like to know about \textless{}country\textgreater{}, can you describe it?                                                                                \\
information seeking             & Imperative                      & Describe \textless{}country\textgreater to me.                                                                                                               \\
information seeking             & Genitive                        & \textless{}country\textgreater{}'s description, give it to me.                                                                                               \\
information seeking             & Relativization                  & I want a description of \textless{}country\textgreater that you know.                                                                                        \\
content generation              & Active                          & I want to generate content about \textless{}country\textgreater and I want you to describe it.                                                               \\
content generation              & Passive                         & A description of \textless{}country\textgreater is desired for content generation, can you describe it?                                                      \\
content generation              & Conjunction                     & I want to generate content about \textless{}country\textgreater{}, and can you describe it to me?                                                            \\
content generation              & Interrogation                   & What is a good description of \textless{}country\textgreater for generating content?                                                                         \\
content generation              & Pronominalization               & I'd like to generate content on \textless{}country\textgreater{}, can you describe it?                                                                       \\
content generation              & Imperative                      & Describe \textless{}country\textgreater to me so I can generate content.                                                                                     \\
content generation              & Genitive                        & \textless{}country\textgreater{}'s description for content generation, give it to me.                                                                        \\
content generation              & Relativization                  & I want a description of \textless{}country\textgreater that you know for content generation.                                                                 \\
\rowcolor[HTML]{CBCEFB} 
Original                        & Original                        & Tell me about \textless{}country\textgreater{}?                                                                                                              \\
information seeking             & Imperative                      & Tell me about \textless{}country\textgreater{}                                                                                                               \\
information seeking             & Active                          & I want to know about \textless{}country\textgreater{}.                                                                                                       \\
information seeking             & Passive                         & How is information about \textless{}country\textgreater gathered?                                                                                            \\
information seeking             & Conjunction                     & I am studying \textless{}country\textgreater and want to know more about it.                                                                                 \\
information seeking             & Interrogation                   & What should I know about \textless{}country\textgreater{}?                                                                                                   \\
information seeking             & Pronominalization               & \textless{}country\textgreater is a place I want to visit, tell me about it.                                                                                 \\
information seeking             & Genitive                        & \textless{}country\textgreater{}'s important information, tell me.                                                                                           \\
information seeking             & Relativization                  & I want information about  \textless{}country\textgreater that is relevant.                                                                                   \\
content generation              & Imperative                      & Tell me about \textless{}country\textgreater{}                                                                                                               \\
content generation              & Active                          & I want to create content about \textless{}country\textgreater{}.                                                                                             \\
content generation              & Passive                         & How can content about \textless{}country\textgreater be generated?                                                                                           \\
content generation              & Conjunction                     & I want to write a story and it is set in \textless{}country\textgreater{}, tell me about it.                                                                 \\
content generation              & Interrogation                   & What kind of content can be created, based on  \textless{}country\textgreater{}?                                                                             \\
content generation              & Pronominalization               & \textless{}country\textgreater is a place I want to write about, tell me about it.                                                                           \\
content generation              & Genitive                        & \textless{}country\textgreater{}'s stories, tell me more.                                                                                                    \\
content generation              & Relativization                  & I want content created, based on \textless{}country\textgreater and it's culture.                                                                            \\
\rowcolor[HTML]{CBCEFB} 
Original                        & Original                        & Write a few paragraphs about the culture of \textless{}country\textgreater{}?                                                                                \\
information\_seeking            & Active                          & I want to learn about the culture of \textless{}country\textgreater{}.                                                                                       \\
information\_seeking            & Passive                         & How is the culture of \textless{}country\textgreater understood?                                                                                             \\
information\_seeking            & Conjunction                     & I am curious about \textless{}country\textgreater and I want to know about its culture.                                                                      \\
information\_seeking            & Interrogation                   & What are some key aspects of the culture of \textless{}country\textgreater{}?                                                                                \\
information\_seeking            & Pronominalization               & \textless{}country\textgreater has a rich culture, what are some of its defining characteristics?                                                            \\
information\_seeking            & Imperative                      & Describe the culture of \textless{}country\textgreater in detail.                                                                                            \\
information\_seeking            & Genitive                        & \textless{}country\textgreater{}'s culture, what are its main components?                                                                                    \\
information\_seeking            & Relativization                  & I'm interested in the culture that is found in \textless{}country\textgreater{}.                                                                             \\
content\_generation             & Active                          & I want to write about the culture of \textless{}country\textgreater{}.                                                                                       \\
content\_generation             & Passive                         & How can the culture of \textless{}country\textgreater be represented?                                                                                        \\
content\_generation             & Conjunction                     & I need to create content and it will be about the culture of \textless{}country\textgreater{}.                                                               \\
content\_generation             & Interrogation                   & What are some interesting points to cover when writing about the culture of \textless{}country\textgreater{}?                                                \\
content\_generation             & Pronominalization               & \textless{}country\textgreater has a vibrant culture, what are some good paragraphs about it?                                                                \\
content\_generation             & Imperative                      & Draft some paragraphs that describe the culture of \textless{}country\textgreater{}.                                                                         \\
content\_generation             & Genitive                        & \textless{}country\textgreater{}'s culture, generate a paragraph about it.                                                                                   \\
content\_generation             & Relativization                  & I am writing about the culture that exists in \textless{}country\textgreater{}.                                                                              \\
\rowcolor[HTML]{CBCEFB} 
{\color[HTML]{333333} Original} & {\color[HTML]{333333} Original} & {\color[HTML]{333333} Can you describe \textless{}country\textgreater{}?}                                                                                    \\
information seeking             & Interrogation                   & Can you describe \textless{}country\textgreater{}?                                                                                                           \\
information seeking             & Active                          & I want you to describe \textless{}country\textgreater{}.                                                                                                     \\
information seeking             & Passive                         & How is \textless{}country\textgreater described?                                                                                                             \\
information seeking             & Conjunction                     & I am curious and want you to describe \textless{}country\textgreater{}.                                                                                      \\
information seeking             & Pronominalization               & \textless{}country\textgreater is interesting, please describe it.                                                                                           \\
information seeking             & Imperative                      & Describe \textless{}country\textgreater{}.                                                                                                                   \\
information seeking             & Genitive                        & \textless{}country\textgreater{}'s description, provide it.                                                                                                  \\
information seeking             & Relativization                  & I'm interested in a description of \textless{}country\textgreater that you can provide.                                                                      \\
content generation              & Interrogation                   & Tell me about the process of describing \textless{}country\textgreater{}.                                                                                    \\
content generation              & Active                          & I want to create a description of \textless{}country\textgreater{}.                                                                                          \\
content generation              & Passive                         & How can descriptions of \textless{}country\textgreater be generated?                                                                                         \\
content generation              & Conjunction                     & Let's describe a country and that country is \textless{}country\textgreater{}.                                                                               \\
content generation              & Pronominalization               & \textless{}country\textgreater needs a description, create one.                                                                                              \\
content generation              & Imperative                      & Outline steps for describing \textless{}country\textgreater{}.                                                                                               \\
content generation              & Genitive                        & \textless{}country\textgreater{}'s description generation, how should I approach it?                                                                         \\
content generation              & Relativization                  & I want to explore the ways that \textless{}country\textgreater can be described.                                                                            
\end{tabular}}
\caption{Table represents the diversified prompts used to capture LLM responses about culture of different countries.}
\label{tab:prompts}
\end{table*}

\newpage
\subsection{Autorater prompt for detecting Facets within LLM responses}
\label{autorater-instructions}
\begin{quote}
\textbf{Objective}: Your goal as an autorater is to analyze the provided text about different countries and identify the presence of specific cultural aspects within the text.\\

\textbf{Task}: For each provided text, you will generate a list of all cultural aspects mentioned.\\

\textbf{Cultural Aspects and Definitions}:

Below is a comprehensive list of cultural aspects we are interested in. For each aspect, a short definition and an example are provided to guide your identification.

\textbf{Fashion/Style}: Refers to the prevalent styles of clothing, accessories, and hairstyles within a given culture.\\
\textit{Definition:} The typical attire, adornment, and hair grooming practices of a society or group.\\
\textit{Example}: ``Many women in Saudi Arabia wear an abaya when in public.''

\textbf{Cuisines}: Encompasses the distinctive culinary traditions, food preparation methods, and typical dishes of a country or region.\\
\textit{Definition}: The characteristic style of cooking practices and traditions, often associated with a specific region or country.\\
\textit{Example}: ``Italian cuisine is famous for its pasta dishes like spaghetti carbonara.''

\textbf{Architecture/Physical Spaces}: Describes the design and construction of buildings and the characteristics of public and private spaces within a culture. This includes markets, houses, religious spaces, schools, parks, neighborhoods, buildings, and landmarks.\\
\textit{Definition}: The art and science of designing and constructing buildings and other physical structures, as well as the characteristics of how a society organizes its living and public environments.\\
\textit{Example}: ``The majestic temples of Angkor Wat showcase ancient Khmer architecture.''

\textbf{Performance and Art}: Covers various forms of artistic expression, including literature, visual arts, performing arts, and traditional storytelling. This includes literature/literary works, literary movements, writers, poets, poetry, art movements, artists, theatre, musical genres, musicians, musical instruments, dance forms, dancers, architectural styles (as an artistic expression), folklore and stories, characters, and movies.\\
\textit{Definition}: The creative expression of human imagination and skill, typically in a visual, auditory, or performing medium, that communicates ideas, emotions, or a worldview.\\
\textit{Example}: ``Brazilian samba music is a vibrant musical genre often accompanied by energetic dance forms.''

\textbf{Religious Rituals}: Refers to the prescribed ceremonies and practices performed within a religious context. This includes rites around death or birth and worship rites.\\
\textit{Definition}: Formalized, symbolic actions or procedures performed for a religious purpose, often at specific times or occasions.\\
\textit{Example}: ``In Hinduism, the 'puja' ritual is performed to worship deities.''

\textbf{Social Practices/Customs}: Encompasses the conventional and generally accepted behaviors, manners, and activities of daily life within a society. This includes rites of passage, behaviors of daily life (e.g., how to eat, sit), and leisure activities.\\
\textit{Definition}: The established ways of behaving or doing something within a society or group, often passed down through generations.\\
\textit{Example}: ``In Japan, it's customary to bow when greeting someone.''

\textbf{Events}: Refers to significant occurrences or celebrations within a culture, both public and private. This includes weddings, funerals, national holidays, birth rituals, and religious holidays.\\
\textit{Definition}: Occasions of significance or celebrations that are part of a cultural tradition or calendar.\\
\textit{Example}: ``Diwali, the festival of lights, is an important Hindu religious holiday.''

\textbf{Sports}: Refers to the games, athletic competitions, and recreational physical activities popular within a culture.\\
\textit{Definition}: Organized recreational or competitive physical activities pursued for entertainment, exercise, or competition.\\
\textit{Example}: ``Cricket is a hugely popular sport in India and many Commonwealth countries.''

\textbf{VNBM (Values, Norms, Beliefs, Morality)}: Represents the fundamental principles, acceptable behaviors, ethical standards, and shared convictions that guide a society. This includes values, norms (acceptable behaviors, gender relationships, class relationships), morality, and beliefs (mythology, religious beliefs, deities).\\
\textit{Definition}: The collective principles, rules, and convictions that define what is considered good, right, and true within a society, and how individuals are expected to behave.\\
\textit{Example}: ``The concept of 'filial piety' (respect for elders) is a deeply ingrained value in many East Asian cultures.''

\textbf{Important Figures}: Refers to notable individuals who have played significant roles in a country's history, politics, or cultural life. This includes political leaders, celebrities, and historic figures.\\
\textit{Definition}: Individuals who hold significant influence or recognition within a society due to their achievements, status, or historical impact.\\
\textit{Example}: ``Nelson Mandela is an important historic figure in South Africa's struggle against apartheid.''

\textbf{History}: Encompasses the past events, narratives, and foundational myths that shape a country's identity and understanding of itself. This includes founding myths, historic events, and folklore and stories.\\
\textit{Definition}: The aggregate of past events, particularly those affecting a country, and the narratives or interpretations of these events that contribute to a collective identity.\\
\textit{Example}: ``The French Revolution is a pivotal historic event that shaped modern France.''

\textbf{Communication}: Describes the various ways in which people interact and convey information, including spoken, written, and non-verbal forms. This includes written language, oral language, non-verbal cues, sign language, and patterns of speech/idioms.\\
\textit{Definition}: The process of conveying information, ideas, or feelings through various verbal and non-verbal means.\\
\textit{Example}: ``The widespread use of metaphors and proverbs is a common pattern of speech in many African oral traditions.''\\

\textbf{Output Format:}

For each text analyzed, your output should be a single list containing all identified cultural aspects. The list should only contain the name of the cultural aspect, as defined above (e.g., ``Fashion/Style'', ``Cuisines'').

Example of Input and Desired Output:

\textbf{Input Text:} ``In Japan, sushi is a very popular dish, and many people enjoy visiting ancient temples. The traditional tea ceremony is a revered social practice, and bowing is a common form of greeting. Sumo wrestling is also a significant sport. Furthermore, the belief in kami (spirits) is deeply ingrained in Shintoism.''

\textbf{Desired Output: }[``Cuisines'', ``Architecture/Physical Spaces'', ``Social Practices/Customs'', ``Sports'', ``VNBM'']

\end{quote}

\end{document}